\documentclass[letterpaper, 10 pt, journal]{ieeeconf}  

\IEEEoverridecommandlockouts  

\usepackage{amsmath} 
\usepackage{amssymb}  

\usepackage{graphicx}
\usepackage{hyperref}
\usepackage{booktabs} 
\usepackage[T1]{fontenc}
\usepackage{newtxtext}  
\usepackage{pifont}
\usepackage{tikz}
\usepackage{xcolor}
\usepackage{adjustbox}
\usepackage{algorithm}
\usepackage{algpseudocode}
\let\labelindent\relax
\usepackage{enumitem}
\usepackage{color,soul,todonotes}
\usepackage{tabularx}
\usepackage{array}
\usepackage{capt-of} 
\usepackage{cuted}

\makeatletter
\newcommand{\IEEEfirstsection}{%
  \@startsection{section}{1}{\z@}%
  {0.4ex plus 0.2ex minus 0.1ex}
  {0.5ex plus 0.2ex minus 0ex}
  {\normalfont\normalsize\centering\scshape}}
\makeatother

\usepackage{comment}

\newif\ifhidden
\hiddentrue   

\newcommand{\hidden}[1]{%
  \ifhidden\else #1\fi
}

\newcommand{\filledstar}{\ding{72}}
\newcommand{\yellowfilledstar}{{\color{yellow}\filledstar}}

\let\oldtextit\textit
\renewcommand{\textit}[1]{\oldtextit{\textcolor{black}{#1}}}
\usepackage{etoolbox}
\everymath{\color{black}}
\everydisplay{}
\BeforeBeginEnvironment{tabular}{\begingroup\color{black}\everymath{}\everydisplay{}}
\AfterEndEnvironment{tabular}{\endgroup}
\BeforeBeginEnvironment{tabular*}{\begingroup\color{black}\everymath{}\everydisplay{}}
\AfterEndEnvironment{tabular*}{\endgroup}

\title{\LARGE \bf See and Switch: Vision-Based Branching for Interactive Robot-Skill Programming}

\author{Petr Vanc,\quad Jan Kristof Behrens, \quad Václav Hlaváč, and \quad Karla Stepanova
}

\begin{document}
\bstctlcite{BSTcontrol}

\maketitle

\begin{abstract}
Programming by demonstration (PbD) makes robot programming accessible to non-experts, but scaling it to real-world variability remains a challenge for current teaching frameworks, especially when a robot must select suitable task variants online from visual input. 
We present \emph{See \& Switch}, an interactive teaching-and-execution framework that represents tasks as graphs of skill parts connected by decision states, enabling conditional branching during replay. Its vision-based \emph{Switcher} uses eye-in-hand images to select the appropriate successor skill part and detect novel situations that require new demonstrations. The framework supports recovery demonstrations during execution through kinesthetic teaching, joystick control, and hand gestures. We evaluate \emph{See \& Switch} on three dexterous manipulation tasks with 8 novice users, collecting approx. 900 real-robot execution rollouts. To isolate visual decision performance from timing errors during decision states, we evaluate the Switcher offline using user-gated decision state windows. In the evaluation within the decision state windows, the method achieves up to $90.6\%$ branch-selection accuracy and detects anomalies with $>90\%$ accuracy in 47 of 79 decision states, demonstrating reliable switching based on visual input for conditional robot-skill programming.
We provide all code and experiment data at 
\href{http://imitrob.ciirc.cvut.cz/publications/seeandswitch}{imitrob.ciirc.cvut.cz/publications/seeandswitch}.


\end{abstract}

\IEEEfirstsection{Introduction}

Teaching robots from human guidance rather than code makes automation more accessible to non-experts. In \textit{Learning from Demonstration} (LfD), a robot derives its control policy by imitating the observed human behavior (e.g., kinesthetic teaching) instead of relying on hand-crafted code or trajectories  \cite{billard2016learning, ARGALL2009469, CalinonIncrementalLearningGestures2007}.
Building on this idea, \textit{Programming by Demonstration} (PbD) enables end-users to teach robots interactively through demonstrations, without writing any code or tuning parameters \cite{Billard2008, eiband2023collaborative, billard2016learning}. 

\begin{figure}
  \centering
  \includegraphics[width=0.49\textwidth]{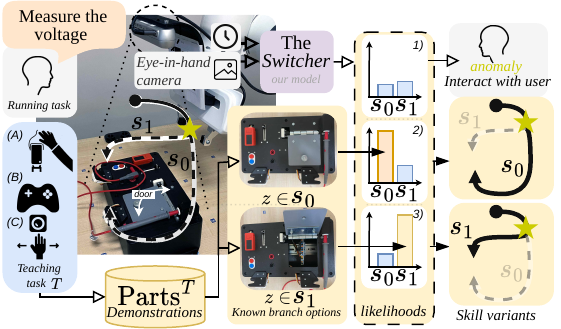}
  \vspace{-2em}
\caption{\textbf{Interactive robot teaching framework.} The user requests to \emph{measure the voltage}. 
During execution, a robotic trajectory is replayed, and the \yellowfilledstar~marks a decision state (DS). At this point, the system may select the most suitable successor skill part $s_0$ (option \textit{2)}) or $s_1$ (option \textit{3)}) or trigger an anomaly (option \textit{1)}) if no previously seen options fit the observation. The Switcher is described in Sec.~\ref{sec:switcher}.}
\vspace{-2.6em}
  \label{fig:introfigure}
\end{figure}

This demonstration-based approach is intuitive, but a single demonstration is only valid in the environment state in which it was given.
Consider an example task on the Robothon taskboard~\cite{robothon} shown in Fig.~\ref{fig:introfigure}: a user teaches a robot to measure a voltage by guiding a probe to the contact point. Replaying the demonstration succeeds when the contact point is accessible, but fails when it is hidden behind
a closed door, since the door has to be open first.  
Handling such variability
demands two capabilities that fixed-replay PbD lacks. First, the robot must recognize from its own observations which of the demonstrated behaviors the current situation requires. Second, when no demonstrated behavior fits, the user must
be able to supply a corrective demonstration on the spot. 
This motivates task representations that can be incrementally extended through human feedback and that encode alternative behaviors as conditional skills.

Graph-based PbD frameworks~\cite{Konidaris2012CTS, eiband2023collaborative, billard2016learning} represent such conditional behaviors using task-graph models with multiple \emph{skill variants}, where a skill variant defines a complete task solution as an end-to-end execution path from task start to completion. Each skill variant is decomposed into \textit{skill parts}, i.e., demonstrated trajectory segments corresponding to alternative environmental states, connected at \textit{decision states} (DS), where execution continues along one of several successor branches (Fig.~\ref{fig:introfigure}). The Collaborative Incremental Programming (CIP) system by Eiband et al.~\cite{eiband2023collaborative} additionally grows this graph incrementally, using an \textit{anomaly detector} to flag execution contexts that are out-of-distribution (OOD), and the user is asked to provide a new demonstration. What remains open is the decision mechanism itself. Existing systems typically rely on low-level internal signals such as force sensing~\cite{eiband2023collaborative}, which reliably capture contact and motion states but are limited to what the robot senses through its own body. However, a closed door produces no force signal until the robot collides with it. Vision-based branching~\cite{Migimatsu2022SymbolicSE} enables contact-free access to environmental context, but still relies on manually specified perceptual features.

We present \emph{See \& Switch}, a teaching-and-execution graph-based PbD framework that grounds the decisions in vision and enables incremental expansion of the task graph via human feedback. A core is an image-based \emph{Switcher} model (Sec.~\ref{sec:switcher}) that evaluates the eye-in-hand camera image at each decision state to select the appropriate successor skill part, or to flag the context as OOD when no demonstrated branch fits. A confirmed anomaly becomes a new conditional decision state and expands the task graph with situation-specific corrective demonstration provided during execution by the user via a modality-agnostic teaching interface (kinesthetic guidance, joystick control, or hand gestures). This enables users to add new skill parts online with minimal effort and without restructuring the existing task representation.  Building on the CIP-style task-graph framework~\cite{eiband2023collaborative}, our contributions are:

\begin{itemize}
\item A DS-local visual \emph{Switcher} that selects among the successor skill parts available at the current decision state and detects out-of-distribution contexts based on the eye-in-hand observations, using frozen foundation-model features. Restricting inference to the local branch set avoids global scene classification and supports operation from sparse user demonstrations.
\item \emph{Incremental task-graph expansion} with automatic insertion of decision states triggered by (visual) anomaly detection~\cite{8798604} or user intervention,  each resolved by a user-guided recovery demonstration that adds a new skill variant without modifying the existing structure and in parallel incrementally updates the Switcher;
\item \emph{Modality-agnostic teaching} that lets users provide new skill variants through kinesthetic guidance, joystick, or hand gestures interchangeably.
\item We evaluate the framework in a user study with 8 users and 3 tasks, comparing teaching modalities in terms of feasibility and teaching efficiency.
\end{itemize}
All source code, interactive visualization tools, experimental videos, and datasets are available at: 
{\footnotesize \href{http://imitrob.ciirc.cvut.cz/publications/seeandswitch}{\texttt{imitrob.ciirc.cvut.cz/publications/seeandswitch}}}.
\vspace{-1em}

\section{Related Work}
\label{sec:sota}

PbD has progressed from single-trajectory replay to graph-structured task models capturing conditional behavior. Neural Task Graph Networks~\cite{Huang2019NeuralTaskGraphs} infer latent task graphs from a single video, without interactive refinement.  CIP~\cite{8798604,eiband2023collaborative} learns task graphs with decision states and recovery branches, triggering anomalies and selecting branches from proprioceptive and force signals. CoBT~\cite{Jain2024CoBT} induces reactive behavior trees from a single demonstration, grounding their conditions in motion-capture object poses. The conditional structure is thus available, but the perceptual decision mechanism is not. Force-based selection requires contact with the discriminating feature, and external trackers presuppose an instrumented, fully observable scene. Eiband et al.~\cite{eiband2023collaborative} explicitly name visual cues, observable before contact, as an open extension. We adopt the user-editable task-graph representation and supply this mechanism with an image-based classifier that operates on context windows of individual decision states.

Execution monitoring detects when a skill no longer fits the observed situation. ConditionNET~\cite{Sliwowski2025ConditionNET} and ILeSiA~\cite{ilesia} learn action preconditions/effects and visual risk scores, respectively, while symbolic pipelines~\cite{Hegemann2022SymbolicFailure,Migimatsu2022SymbolicSE} ground manually defined predicates in multimodal signals. These monitors report only whether execution may proceed, but the response to failure remains fixed---utilizing pre-planned recovery branches~\cite{Sliwowski2025ConditionNET} or separately learned recovery motions that assume the anomaly is already identified~\cite{Wu2021Learning}. Our Switcher extends ILeSiA's binary risk score to a multiclass decision at each decision state where a single embedding space both flags OOD views and selects among competing skill parts, tightly coupling branch selection and anomaly detection. A confirmed anomaly is resolved by the user demonstrating a new branch during execution, which becomes visually selectable in subsequent executions.

Vision foundation models increasingly serve as frozen state encoders for manipulation. DINOBot~\cite{DiPalo2024DINOBot} uses DINO~\cite{dino} features for retrieval-based one-shot imitation and pixel-level alignment to novel objects. Diffusion-PbD~\cite{Murray2024DiffusionPbD} transfers demonstration waypoints to novel scenes via diffusion features. They generalize how a single demonstrated behavior executes in a new scene, but do not decide which of several demonstrated behaviors the scene requires, nor detect that none fits. We use the same frozen DINO features within a decision-state-specific classifier to support few-shot branch selection and anomaly detection.


\section{Method}
\label{sec:method}

We present a modality-agnostic system for incrementally programming \textit{conditional skills} (tasks with multiple skill variants) from user demonstrations. The approach integrates online execution monitoring and mechanisms for teaching recovery behaviors, formalized in the robot policy described in Sec.~\ref{sec:policy}. Our design retains the decision-state (DS) logic and anomaly-triggered insertion rule of CIP~\cite{eiband2023collaborative}. 
We introduce (\emph{i}) an \textit{input-modality-abstraction layer} that unifies teaching interfaces (Sec.~\ref{sec:teaching}), and (\emph{ii}) the \textit{ Switcher: vision-based} observation channel (eye-in-hand) capable of online detecting anomalies and selecting the skill continuation at each timestep, Sec.~\ref{sec:switcher}. 
Fig.~\ref{fig:system_diagram} gives an overview of the resulting teaching-and-execution framework.

\subsection{Problem Formulation and Task Representation}
\label{sec:problem}

We represent a task by a task-graph $G$ consisting of a set of \textit{skill parts} $\text{Parts}^T=\{s_{1},\dots,s_{M}\}$ connected at \textit{decision states}. Each skill part $s_i$ is created based on a single demonstration and optionally refined with $R \in \mathbb{N}$ execution rollouts of this demonstration. The $i$-th skill part $s_i$ is a tuple:

$$s_{(i)}= \left(\left( p^t_{(i),r}\,, \:{g}^t_{(i),r}\,, \:z^t_{(i),r}\right)_{t=0}^{N_{(i)}} \,, \:K_{(i)} \right)_{r=0}^{R}\,,$$
where $p^t_{(i),r} \in \mathbb{R}^7$ is observed robot pose toward the base, $g^t_{(i),r} \in \mathbb{R}$ is gripper state (open/close), and $z^t_{(i),r} \in \mathbb{R}^{224 \times 224}$ is observed eye-in-hand image at time $t$ in the rollout $r$. $N_{(i)}$ is the number of timesteps of the $i$-th skill part. 
$R$ execution rollouts for $i$-th skill part are used for training the Switcher, Sec.~\ref{sec:switcher}. 
The offset $K_{(i)} \in \mathbb{N}_0$ is the timestep at which $s_{(i)}$ begins, i.e. skill part is only temporally valid at $t \geq K_{(i)}$. The replay controller for $s_i$ is the time-indexed reference $\pi_{(i)}(t) := p^t_{(i),0}$.

Skill parts are connected at \textit{decision states}. Each decision state (DS) $d \in D$ is located at a branching timestep $t_d$ on the task timeline and is defined by (i) a \textit{DS context window} $W_d = \{t_d, \dots, t_d + e\}$ of length $e$, the interval over which observations inform the decision, and (ii) \textit{permitted successor skill parts} $M^U_d \subseteq \text{Parts}^T$ that may be executed from $d$. 

Following the~\cite{eiband2023collaborative}, a \textit{task graph} $G=(D, \mathrm{Parts}^T)$ consists of decision states as branching nodes and skill parts $\mathrm{Parts}^T$ as directed edges between them (Fig.~\ref{fig:taskgraph}). A \textit{skill variant} is one complete path through $G$ from task start to completion. Because each DS is reached along a single incoming skill part $s_{(i)}$, we index the set of permitted successor skill parts to which we can switch from skill part $s_{(i)}$ at the context window $W_d$ of this DS by that part, writing $M^U_{(i)}$ for $M^U_d$. Restricting decisions to $W_d$ and $M^U_{(i)}$ keeps them local, i.e., the system never classifies the global scene, only discriminates among the few alternatives demonstrated at this point of the task. For every new task, the task graph is not given in advance: teaching starts from a single demonstration, $\text{Parts}^T = \{s_{(0)}\}$, $D = \emptyset$, and the graph is incrementally expanded with new alternatives provided during execution.

Our method deals with three coupled sub-problems, which are discussed in the remainder of this section: (i) \textbf{Branch selection.} At a DS $d$, given images $z^t$, $t \in W_d$, select the successor skill part $i_p \in M^U_d$ matching the observed environment state. (ii) \textbf{Anomaly detection.} Decide, for the current observation, whether the scene lies outside all demonstrated contexts. (iii) \textbf{Incremental expansion.} When an anomaly is confirmed, acquire a recovery demonstration during execution, insert it as a new skill part, and update the decision mechanism, without invalidating previously demonstrated behavior.
\vspace{-1.5em}
\subsection{Robot execution policy}
\label{sec:policy}
\vspace{-1em}

\begin{algorithm}[ht!]
\caption{Robot policy $\Pi^T$ for execution of task $T$.}
\label{alg:inf_exec}
{\scriptsize
\begin{algorithmic}[1]
\State $t\gets0$ \Comment{{\scriptsize Initialize execution timestep index $t\in\mathbb{N}_0$}}
\State {\small $(p_{(0),0}^t, g_{(0),0}^t,z_{(0),0}^t, K=0) \gets s_{(i=0),r=0}^t$} \Comment{{\scriptsize Load initial skill part}}
\State $\mathcal{S}\gets \text{train}(Z^{U})$ \Comment{{\scriptsize Train the Switcher on image set, Sec.~\ref{sec:switcher}}}
\State $N_{(i)} \gets |p_{(i)=0, r=0}|$ \Comment{{\scriptsize Get trajectory length}}
\While{$t < N_{(i)}$}
\State $r \gets \text{New execution rollout}$
\State $z^t_{(i),r}, p^t_{(i),r} \gets \text{sense()}$ \Comment{{\scriptsize Observe image and robot proprioception}}
\State $i_{\text{p}}, a_{\text{p}} \;\gets\; \mathcal{S}(z^t_{(i),r},\, t)\,$ \Comment{{\scriptsize The Switcher prediction, Sec.~\ref{sec:switcher}}}
\If{$a_{\text{p}}$ is True}
    \State Anomaly event: (1) Branch or (2) DAggr \Comment{{\scriptsize Sec.~\ref{sec:anomalyevent}}}
\EndIf
\If{$i^p \neq i^{t-1}$, i.e., does target skill part change}
    \State {\small $s_{(i),r}^t \gets (p_{(i),r}^t, g_{(i),r}^t,z_{(i),r}^t, K)$} \Comment{{\scriptsize Save execution rollout}}
    \State {\small $(p_{(i_{\text{p}}),0}^t, g_{(i_{\text{p}}),0}^t,z_{(i_{\text{p}}),0}^{t}, K) \gets s_{(i_{\text{p}}),0}^t$} \Comment{{\scriptsize Load new skill part}}
    \State $N_{(i)} \gets |p_{(i), r=0}|$ \Comment{{\scriptsize Get trajectory length}}
    \State $t \gets 0$ \Comment{{\scriptsize We reset timestep, while knowing offset $K>0$}}
\EndIf
\State {\small $p^t_{(i),0}, g^t_{(i),0} \gets \pi_{p_{(i=i_{\text{p}}),0}}(t)$} \Comment{{\scriptsize Set reference pose, gripper, Sec.~\ref{sec:policy}}}
\If{$||p^{\text{o}} - p^t_{(i),0}|| < \epsilon$, i.e., is near}
    \State $i^t \gets i^p$ \Comment{{\scriptsize save branch prediction}}
    \State $t \gets t + 1$ \Comment{{\scriptsize increment timestep}}
\EndIf
\EndWhile
\State $s_{(i),r=R}^t \gets (p_{(i),R}^t, g_{(i),R}^t,z_{(i),R}^t)$ \Comment{{\scriptsize Save skill execution rollout}}
\end{algorithmic}
}
\end{algorithm}

\begin{figure}[]
    \centering
    \vspace{0.5em}
    \includegraphics[width=0.5\linewidth]{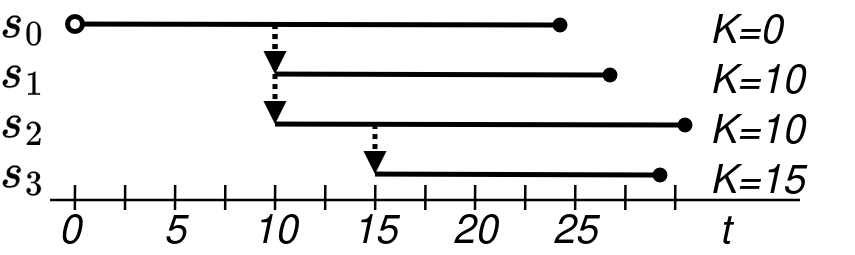}
    \caption{\textbf{Task-graph example.} Four skill parts ($s_{0,1,2,3}$) form four distinct skill variants. Each skill part has an offset $K_{(i)}$ and terminates at different time steps $t$. Decision state (DS) windows are located around $t=10$ and $t=15$. The task-graph grows online through branching, while rollouts are aggregated for Switcher updates.}
    \vspace{-1em}\label{fig:taskgraph}
\end{figure}

A task execution consists of one or more skill parts connected through decision states where switching can happen (Fig.~\ref{fig:taskgraph}). Each execution starts with the initial skill part $s_{(0)}$ ($K_{(0)} = 0$).
Algorithm~\ref{alg:inf_exec} summarizes the overall execution logic and Fig.~\ref{fig:system_diagram} visualizes it. The robot policy $\Pi^T$ combines time-indexed trajectory replay, where the robot tracks the attractor pose $\pi_{(i)}(t - K_{(i)})$ of the active skill part $s_{(i)}$, with a per-timestep decision whether an anomaly has happened ($a_p = \text{True}$, line 8) and whether to switch to a new skill part ($i^p \neq i^{t-1}$, line 12-16). 

While trajectory replay follows standard PbD practice, the key novelty lies in the per-timestep decision logic driven by \emph{the Switcher} $\mathcal{S}$: $(i_p, a_p) \gets \mathcal{S}(z^t, t)$ (Alg.~\ref{alg:inf_exec}, line 8), described in Sec.~\ref{sec:switcher}. Given the image $z_{(i),r}^t$ observed at time $t$, the Switcher predicts the successor skill part $s_{i_p} \in M^U_{(i)}$ and an anomaly flag $a_p \in {0,1}$, indicating whether an anomaly was detected. 
Through these predictions, the Switcher determines the transitions in the task graph by continuously evaluating visual observations during execution.
If $a_p = 0$ and the active skill part changes ($i^t \neq i^{t-1}$, line 12), execution traverses to the successor $s_{(i_p)}$ and replay continues from the reference $\pi_{(i_p)}(t - K_{(i_p)})$ on the shared task timeline (lines 13--17). If $a_p = 1$, no successor is selected and the anomaly-handling logic is invoked (lines 9--11), see also Sec.~\ref{sec:anomalyevent}. When execution has no switches, it remains fully contained within the initial skill part (Fig.~\ref{fig:taskgraph}). The executed skill variant is thus the sequence of skill parts ($s_{(0)},s_{(i)},\dots$) selected online by $\mathcal{S}$ based on the image stream $z^t$.


\vspace{-0.5em}
\subsection{Interactive Incremental Teaching and Recovery}
\label{sec:teaching}

\begin{figure}[t]
  \centering
  \vspace{0.4em}
  \includegraphics[width=0.99\linewidth]{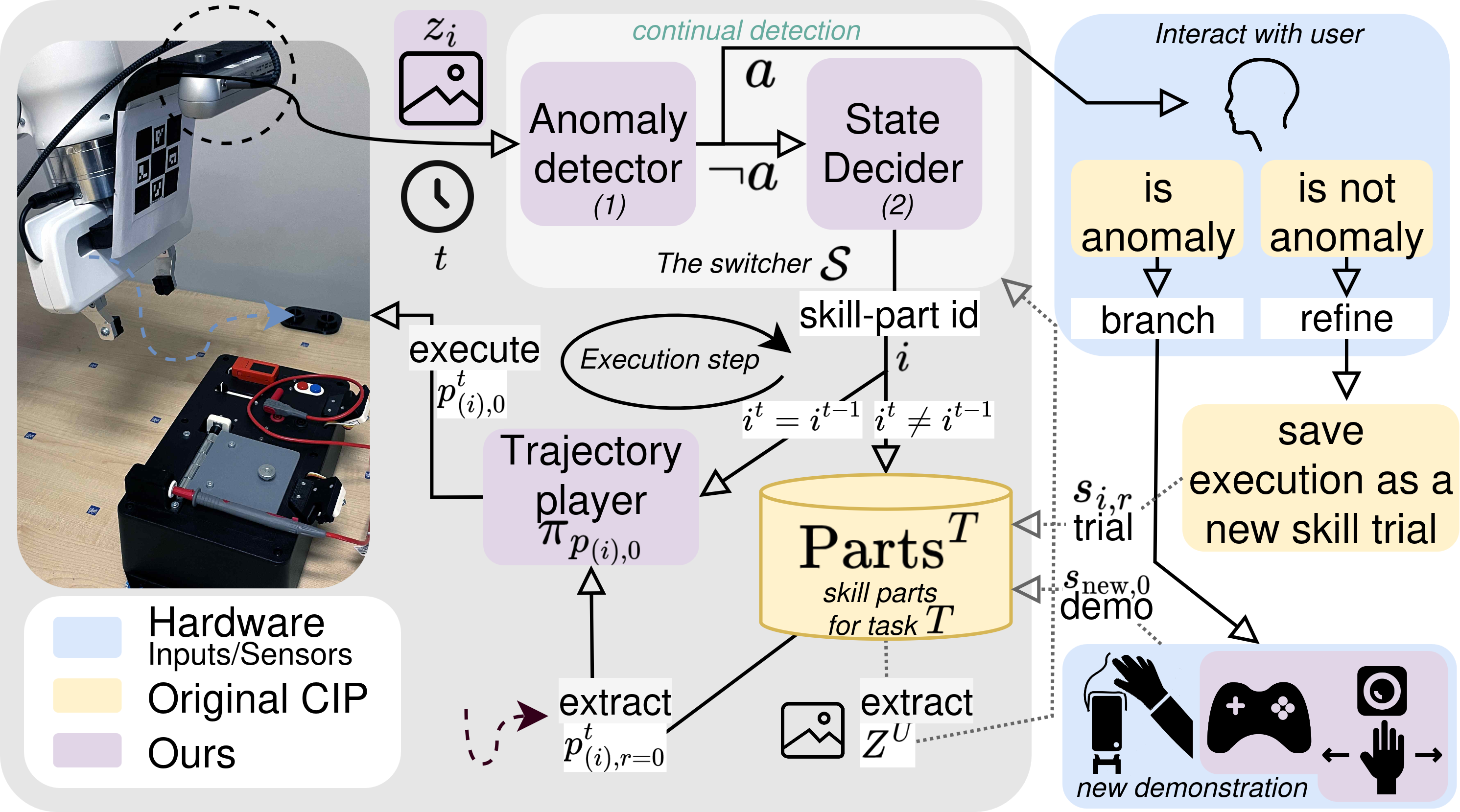}
  \vspace{-2em}
  \caption{\textbf{Interactive robot teaching \& execution framework.} Unchanged CIP core: DS logic where user verifies anomaly $a$, and insertion rule (branch). New components: (1) a modality-agnostic input layer (gestures/joystick/kinesthetic) that maps human intent to robot controls, (2) an optional eye-in-hand vision channel.
  $Z^U$ is a subset of images used for training the Switcher, defined in Sec.~\ref{sec:switcher}. When the skill part $i^t$ is different from the previous ($i^t \neq i^{t-1}$), we load and extract a new skill part trajectory from the library $\text{Parts}^T$.
  \vspace{-2em}
  }
  \label{fig:system_diagram}
\end{figure}

The user first provides an \textit{initial demonstration} of task $T$ using any supported modality (gestures, joystick, or kinesthetic teaching), which defines the initial skill part $s_{(0)}$ and initializes the task representation as $G^T = (\emptyset, \{s_{(0)}\})$ (Sec.~\ref{sec:problem}). The resulting demonstration defines a valid motion trajectory for those specific states of the environment (e.g., probe location).
During execution, the system monitors the image stream $z^t$ at decision points to detect novel (OOD) behavior, the \textit{anomaly}. When such a case is detected, the user may provide an alternative demonstration that adds a new skill variant as a conditional branch, incrementally expanding $G^T$. The accumulated variants enable increasingly autonomous execution across changing conditions. See Fig.~\ref{fig:system_diagram} for the human-robot interaction workflow overview.

\subsubsection{Anomaly event}
\label{sec:anomalyevent}
\begin{figure}
    \centering
    \includegraphics[width=0.9\linewidth]{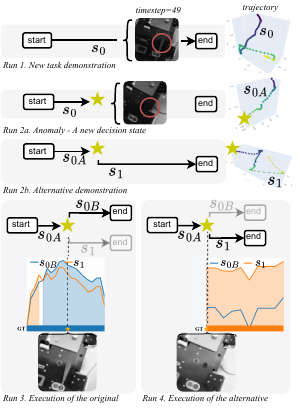}
    \vspace{-1em}
    \caption{\textbf{Teaching a ``peg pick'' task with four separate runs.} \yellowfilledstar~is a decision state. You can see eye-in-hand image at DS (timestep $t=49$) and peg visible/absent. (bottom) You can see the likelihoods for two test runs around DS.}
    \vspace{-1em}
    \label{fig:placeholder}

\end{figure}

An anomaly detected at time $t$ suggests a new potential DS based on visual observations (see Fig.~\ref{fig:placeholder}).
At this point, the user determines whether the observed situation requires a new recovery behavior or corresponds to a known execution context. If the anomaly is confirmed, a new branch is created by demonstrating an appropriate skill part (\textit{Branch}, Sec.:~\ref{sec:branchoption}). Otherwise, execution proceeds without branching and the newly observed data are used to update the Switcher training set without modifying the task-graph structure (\textit{Refine}, Sec.~\ref{sec:refineoption}).

\paragraph{Branch (recovery behavior)} \label{sec:branchoption}
The user demonstrates a new skill part $s_{\text{new}}$ suitable for the current context, i.e., introducing a new branch in the task graph.
We represent the insertion of a DS as a split of the original skill part $s_{\text{old}}$ into two segments, $s_{\text{oldA}}$ and $s_{\text{oldB}}$, with the DS located between them (see Fig.~\ref{fig:placeholder}). A direct edge from the DS to the $s_{\text{new}}$ is added. 
The permitted successor set of the new DS is initialized as $M^U_d \gets \{ s_{\text{oldB}} \}$ and extended with each skill part demonstrated from this DS, $M^U_d \gets M^U_d \cup \{ s_{\text{new}} \}$.
The Switcher $\mathcal{S}$ is retrained to distinguish between the existing successor $s_{\text{oldB}}$ and the newly introduced successor $s_{\text{new}}$, thereby updating the DS-local classifier.

\paragraph{Refine via Data Aggregation (DAggr)} \label{sec:refineoption}
If the user approves continuation (e.g., thumbs-up gesture), execution continues. Captured data $s_{(i),r}$ are saved as an additional execution rollout and used to update the Switcher, i.e., a data-aggregation update that grows the classifier's training set while preserving the task-graph structure.

\subsubsection{Input–Modality Layer (Modality–Agnostic API)}
\label{sec:input}
The input–modality layer provides the interface through which the human guides the robot during programming and execution. 
Input channels (\textit{hand gestures}, \textit{joystick/keyboard}, \textit{kinesthetic teaching}) are unified into a common set of abstract commands, allowing teaching and decision-making components to operate independently of the interaction modality.
We distinguish two primary roles of human input, each mapped to abstract commands as summarized in Table~\ref{tab:modalitymapping}:
\begin{enumerate}
  \item \textit{Demonstration.} The user provides an initial or corrective demonstration, defining how the robot should perform a skill in a given context. This may be the initial trajectory of a new task or a recovery behavior appended at a DS.
  \item \textit{Execution controls.} During skill replay, the user may influence execution by issuing commands such as pause (to inspect for an anomaly) or continue (when the system raises 'Anomaly', and it is a false alarm).
  \hidden{\item \textit{Manual skill part selection.} 
  The user selects the target skill part from finite choices. We consider that each skill part is represented with a user‑intelligible via‑point, Sec.~\ref{sec:intentdisambiguatingviapoints}. The user selects a target via-point either by pointing with a deictic gesture, correcting the robot with kinesthetic nudge or by moving with a joystick D-pad.
  }
\end{enumerate}

\begin{table}[t]
\vspace{0.5em}
\begin{adjustbox}{width=\columnwidth,center}
\centering
\begin{tabular}{p{0.11\linewidth} p{0.33\linewidth} p{0.1\linewidth} p{0.15\linewidth} p{0.15\linewidth}}
\toprule
\textbf{Modality} & \textbf{Demonstration} \newline [target poses] & \textbf{Anomaly} [boolean] & \textbf{Approve} [boolean] & \textbf{Gripper} [open/close] \\ 
\midrule
Kinesthetic & Manual guidance with gravity compensation & ``Pilot'' button & ``Check'' button & ``Cross'' button \\ 
\midrule
Joystick & L stick (x/y), R stick (roll/pitch), \texttt{LB/RB} (z) & \texttt{Y} btn & \texttt{X} btn & \texttt{A/B} btn
\\ 
\midrule
Hand \newline gestures & 6-DoF hand teleop & ``Stop'' gesture & ``Thumbs-up'' gesture & Hand open/close \\ 
\bottomrule
\end{tabular}
\end{adjustbox}
\vspace{-1em}
\caption{Mapping of input modalities to the common command API.}
\vspace{-3em}
\label{tab:modalitymapping}
\end{table}

\vspace{-0.5em}
\subsection{The Switcher (Observation-based Branch Classifier)}
\label{sec:switcher}
The core contribution of our system is the Switcher $\mathcal{S}$: a
shared embedding underlies both anomaly detection and branch selection. At each timestep, given the observed image $z^t$, it either recognizes a known context and selects the corresponding successor skill part, or flags $z^t$ as novel, triggering  an anomaly, see Fig.~\ref{fig:system_diagram}. If the user approves the anomaly, then they demonstrate a new skill part (new branch), which inserts a DS and updates $\mathcal{S}$ immediately, see Fig.~\ref{fig:placeholder}. The Switcher $\mathcal{S}$ is trained on images $Z^U \subset Z$ collected from all skill parts $\text{Parts}^T$ available for the task $T$ and 
has two components built on the shared embedding: 1) an anomaly detector,
and 2) a lightweight, per-DS branch classifier:

\subsubsection{Anomaly detector}

Following ILeSiA~\cite{ilesia}, we monitor a per-timestep risk score that estimates whether a new image is OOD relative to the training set; unlike~\cite{ilesia}, the score is computed in the same frozen DINO embedding space used for branch selection (Sec.~\ref{sec:switcher_methods}), which lets the system decide whether an anomalous view corresponds to a known alternative or a novel context.
We want the anomaly detector to flag changes such as: (1) a new object appears/disappears, (2) the pose or state of the object changes, and (3) the camera view changes, meaning the robot is not respecting the predefined trajectory path. In practice, it should alert us when the probe is misplaced or an obstacle is blocking access. 
However, when anomalies are user-triggered, visual cues may be insufficiently distinctive, preventing reliable association with existing states and transitions.






\subsubsection{Branch Classifier}
\label{sec:stateestimator}

For each new DS created, we train a dedicated Branch Classifier model using images $Z_d = Z^{W_d}_{M^U_d} \subset Z$ from the permitted (temporally valid) successor skill parts 
$M^U_{d}$ observed at the DS window $W_d$ (Sec.~\ref{sec:problem}). See Sec.~\ref{sec:branchoption} for details on how $M^U_d$ grows incrementally with new skill parts demonstrated at the given DS. 
Restricting switching to $M^U_{\mathrm{d}}$ prevents ambiguity arising from visually similar states that correspond to different, unobservable environmental conditions. 
For example, if the door is outside the camera view, 
 the image cannot reveal whether it is open or closed, and is therefore equally consistent with the door-open and door-closed skill parts. 

\subsubsection{Proposed DINO-based Implementation}
\label{sec:switcher_methods}

To implement DS-local branch selection in the Switcher, we use self-supervised DINO vision transformers~\cite{dino} as frozen feature extractors. Given an eye-in-hand image $z$, the backbone outputs a set of patch embeddings $\{x_k\}_{k=1}^{n_p}$, $x_k\in\mathbb{R}^{d}$. We convert these patch embeddings into a class prediction over competing skill parts using one of the following strategies:

\paragraph{Prototype inference (mean)}
We form an image representation either by mean pooling (\texttt{mean}) or by concatenating patch embeddings (\texttt{concat}). For each class $c_i$, we compute a prototype by averaging the training representations of that class. At inference time, we assign the query image to the class with maximum cosine similarity to its prototype.

\paragraph{Multiple-instance learning (MIL)}
We treat patch embeddings as a set of instances and learn an attention-based MIL head that aggregates $\{x_k\}$ into a class prediction.

\paragraph{Attention-gated features (attn)}
To emphasize informative regions, we use the last-layer \texttt{[CLS]}-to-patch self-attention weights as patch importance scores. We apply either \emph{hard} gating (discard low-attention patches) or \emph{soft} gating (down-weight them), and reduce multiple attention heads by mean or max before aggregation.

\paragraph{Anomaly detection}
We use the same DINO embedding space for novelty detection by computing an anomaly score from the query image's similarity to the DS training set (e.g., based on the maximum cosine similarity to known classes) and flagging an anomaly when the score drops below a per-class threshold set at the \texttt{percentile\_keep} quantile of that class's training-score distribution, \texttt{percentile\_keep} $\,\in(0,1]$).

\section{Experimental Setup}
\label{sec:experimental-setup}

We evaluate the proposed system on a dataset collected in our user study where users taught robot manipulation tasks on the Robothon Electronic Task Board~\cite{robothon} (Fig. \ref{fig:introfigure}, bottom left). Our experimental setup includes a Franka Robotics Panda robot with an eye-in-hand RealSense D455 camera (Fig. \ref{fig:introfigure}) and Leap Motion sensor for hand tracking~\cite{Weichert_Bachmann_Rudak_Fisseler_2013}.

\subsection{Robothon Manipulation Skills}
\label{sec:scenarios} 

\begin{figure}
    \centering
    \vspace{0.5em}
    \includegraphics[width=0.7\linewidth]{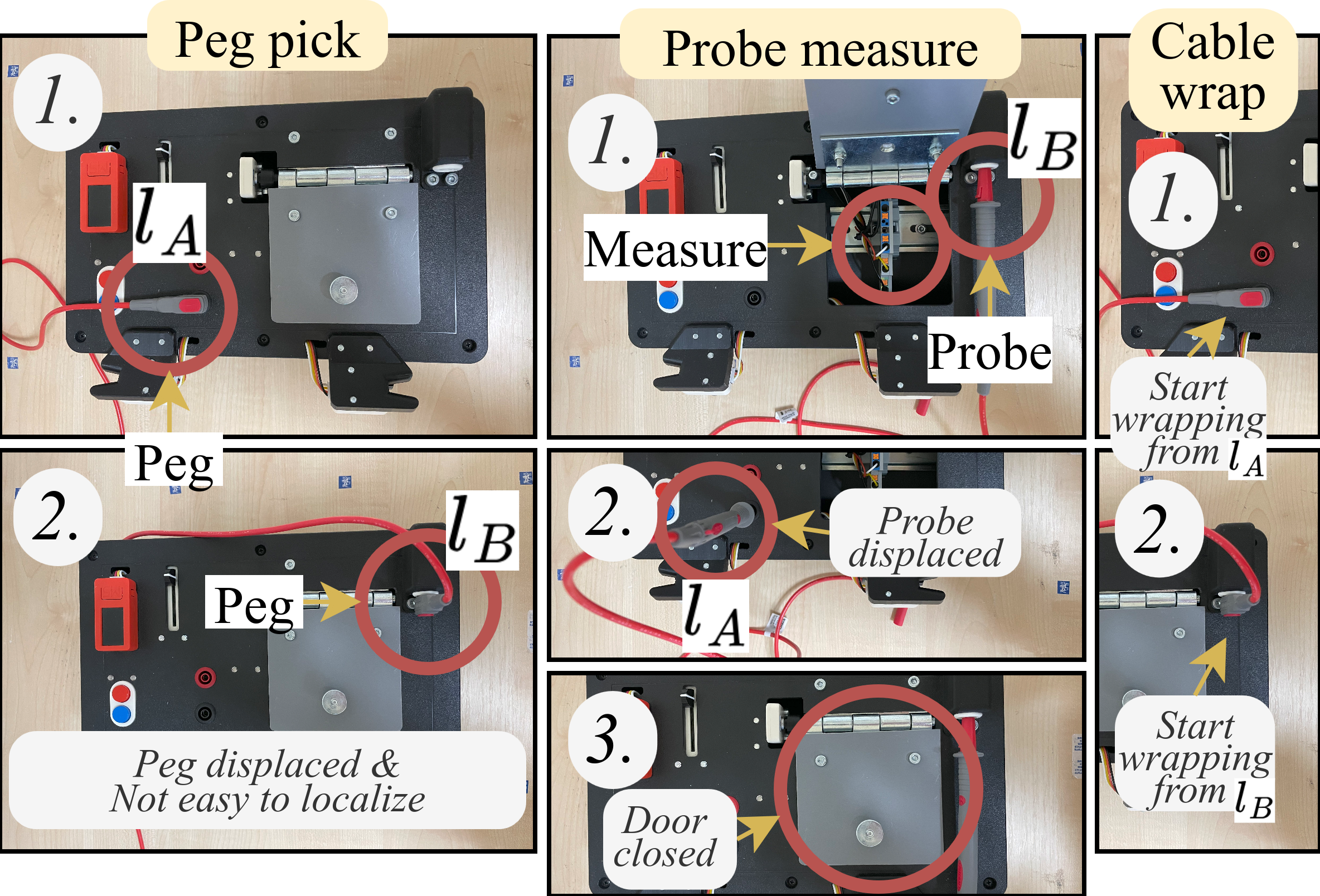}
    \vspace{-1em}
    \caption{\textbf{Starting states of the environment for three considered tasks.}}
    \vspace{0.5em}
    \label{fig:starting_conditions}
    \vspace{-2em}
\end{figure}

We introduce three manipulation tasks on the taskboard: (1)~Peg~pick, (2)~Probe~measure, and (3)~Cable~wrap. The Scenarios are detailed in Fig.~\ref{fig:starting_conditions}.
For each task, we vary the initial state in a way that requires a different type of manipulation skill. This can be, for example, a different initial pose for the peg ($l_A$, $l_B$ in Fig.~\ref{fig:starting_conditions}) or a closed door over the probe measurement point.
In our experiments, mostly novice users teach/program these tasks to \emph{(i)} evaluate the system usability, and at the same time \emph{(ii)} collect the evaluation dataset $\mathcal{D}$.

\subsection{User study dataset configuration}
\label{sec:datasets} 

We collect the dataset $\mathcal{D} = (\mathcal{D}_{demo},\mathcal{R})$ during the user study to evaluate the proposed Switcher (Sec.~\ref{sec:switcher}) for two objectives: (i) target skill-part estimation (branch selection) and (ii) anomaly detection. $\mathcal{D}_{demo}$ is the set of demonstrations and $\mathcal{R}$ the set of rollouts. Each participant was first introduced to the system terminology and task setup (Fig.~\ref{fig:starting_conditions}). To build intuition for the single manipulator and demonstration workflow, we asked participants to first perform the tasks single-handedly.
The users were then asked to perform the task using the given teaching method once. The task is reset to another variation, rendering the original demonstration unsuccessful. The user is asked to flag an anomaly as soon as the culprit is visible in the camera image. From that state on, the user demonstrates the remaining task parts (up to 3 variants) till the task end.

The dataset covers the $3$ manipulation tasks defined in Sec.~\ref{sec:scenarios}.
The study involved $8$ participants (age 24--40). Seven had no prior experience with gesture teleoperation, and five had no prior experience with kinesthetic teaching. The total number of demonstrations is
\begin{equation}
    |\mathcal{D}_{\mathrm{demo}}|
    =
    \underbrace{U}_{8~\mathrm{users}}
    \times
    \underbrace{Q}_{3~\mathrm{modalities}}
    \times
    \underbrace{
        \left(
        V_{\mathrm{peg} + \mathrm{probe} + \mathrm{wrap}}
        \right)
    }_{\substack{2+3+2=7\\\mathrm{skill~variants}}} \\
    = 168.
\end{equation}
\vspace{-1.5em}

Participants typically recorded one demonstration per task and modality. Each successful demonstration defines one skill variant. In approximately $3\%$ of cases, participants repeated the demonstration, replacing the previous attempt. Each demonstration was replayed and evaluated at least three times. Approximately $4\%$ of the demonstrations were considered unsafe, retaining a single recorded rollout, and are reported as unsuccessful in Tab.~\ref{tab:userstudy}. In total, the dataset contains approximately $900$ execution rollouts.

Because users inserted decision states at different points during task execution, each DS may contain either two or three candidate successor variants. The branching timesteps $t_d$ of all inserted decision states are clustered into larger temporal segments by merging overlapping DS windows using a union-of-intervals procedure~\cite{cormen2009introduction} (window length $e = 10$). This resulted in $86$ independent decision-state-window datasets. Each dataset corresponds to one DS window and contains training samples extracted from demonstrations and test samples extracted from execution rollouts.

We treat these DS-window datasets independently: for each DS-window dataset, a separate model is trained and evaluated only on the corresponding test data, and reported accuracies average the per-DS results. 
From the $86$ collected datasets, we excluded $7$ datasets affected by acquisition or setup failures, including frozen camera streams and incorrectly configured scenes. The resulting classification dataset therefore contains $79$ valid datasets including $29$ datasets containing visual distractors which represent valid but more challenging test conditions.

We evaluate branching performance at DS context windows, as defined in Sec.~\ref{sec:problem}, and report results for two objectives: (i) branch selection and (ii) anomaly detection. 
We organize $\mathcal{D}^{79}$ into two labelings. In $\mathcal{D}^{79}_{\mathrm{classification}}$, samples are labeled according to their corresponding skill part, enabling multiclass branch selection. 
In $\mathcal{D}^{79}_{\mathrm{anomaly}}$, samples from the root skill part (the skill part along which the DS is reached) define the in-distribution set, while samples from all other skill parts are labeled as unfamiliar, i.e., out-of-distribution (OOD), for anomaly evaluation.

\subsubsection{Success metric}
\label{sec:replay_success_metric}

The user demonstrations $\text{Parts}^T$ define reproducible task policies $\Pi^T$. We define \textit{Decision success} as when the correct skill variant is selected, and \textit{Task success} if it achieves the task-specific goal: (i) \emph{Peg pick:} grasp the peg (from either location) and drop it into the bowl; (ii) \emph{Probe measure:} touch the measurement point with the probe and then place it into the bowl (including opening the door if it blocks access); (iii) \emph{Cable wrap:} produce at least two complete loops of wrapped cable, regardless of the initial cable-end configuration.

\subsubsection{System-gated vs.\ user-gated anomaly detection}

As described in Sec.~\ref{sec:switcher}, the proposed framework supports both \emph{automatic} DS insertion via visual anomaly detection and \emph{manual} user triggering.
In deployment, both are enabled to maximize robustness. In the user study, however, we intentionally used \emph{user-gated} DS timing. This choice isolates the evaluation of the Switcher and the teaching workflow from temporal errors in automatic triggering: early, late, or missed anomaly detections would otherwise confound branch-selection accuracy and obscure modality comparisons. User-gating also yields consistent DS alignment and reliable labels in the collected dataset, while improving study safety and execution stability by keeping the participant in control. Automatic anomaly detection is evaluated separately offline on the collected data (Sec.~\ref{sec:exp1}).

\subsection{Switcher baselines and configurations}
\label{sec:baselines}

We compare three families of Switcher implementations: (i) classical keypoint matching (SIFT), (ii) an autoencoder + Gaussian Process (AEGP) baseline~\cite{ilesia}, and (iii) DINO-based branch selection (Sec.~\ref{sec:switcher_methods}). For DINO, we evaluate \texttt{DINOv2} and \texttt{DINOv3}~\cite{dinov2,simeoni2025dinov3} with \texttt{small} ($\sim$22M) and \texttt{large} ($\sim$300M) backbones.

\subsubsection{SIFT matching (baseline)}
We extract SIFT/ORB features~\cite{sift} for each training image and store them per class. At inference time, we match a query image to each class template and score the match via a robust homography estimate using MAGSAC++~\cite{magsacpp}.

\subsubsection{Autoencoder--Gaussian Process (AEGP, baseline)}
Following~\cite{ilesia}, we train an autoencoder on DS-window images and use its latent codes as inputs to a GP classifier. Since our branch selection is multiclass (\textit{Multiclass AEGP}), we implement a variational multitask GP with inducing points (one latent function per class).

\subsubsection{Proposed method (DINO-based) configurations}
For the MIL head, we use 128 hidden units, dropout 0.1, and Adam~\cite{adam} (lr $7\times10^{-5}$, weight decay $10^{-3}$), trained for 1000 epochs with cross-entropy loss. For attention-gated features, we report both hard/soft gating, mean/max head reduction, and an attention keep fraction \texttt{attn\_keep}$\,\in(0,1]$ (e.g., \texttt{attn\_keep}=0.2 keeps the top 20\% patches).

\section{Experiments}

We evaluate the framework in three parts: a controlled scalability experiment with an increasing number of Switcher classes~(Sec.~\ref{sec:growing_labels}), an offline Switcher evaluation on the user-study dataset~(Sec.~\ref{sec:exp1}), and observational findings from the study itself---task success and teaching effort across modalities~(Sec.~\ref{sec:exp_userstudy}).
Materials are available on the project website\footnote{\label{websitefootnote} 
\href{http://imitrob.ciirc.cvut.cz/publications/seeandswitch}{imitrob.ciirc.cvut.cz/publications/seeandswitch}.}.
The best-performing \texttt{DINOv2 small attn} Switcher trains in approximately $25$ ms per DS-window dataset and runs inference below $4$ ms per query image, enabling asynchronous updates during real-time execution.

\vspace{-0.5em}
\subsection{Controlled experiment: Model reliability}
\label{sec:growing_labels}

\begin{figure}
    \centering
    \includegraphics[width=0.8\linewidth]{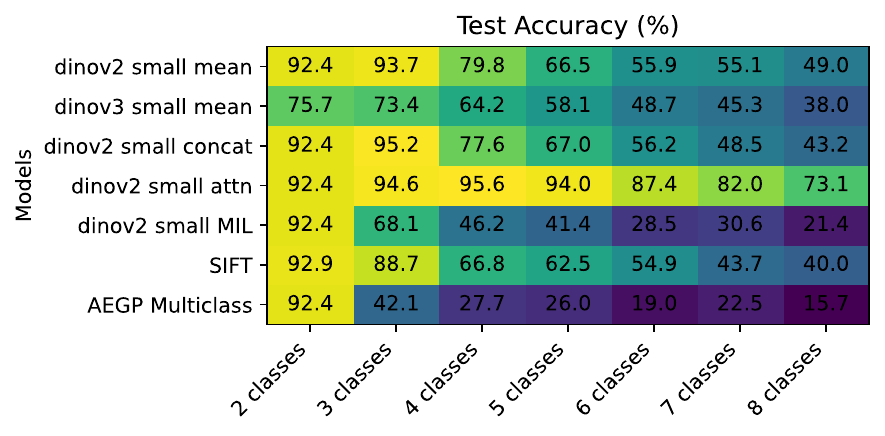}
    \includegraphics[width=0.99\linewidth]{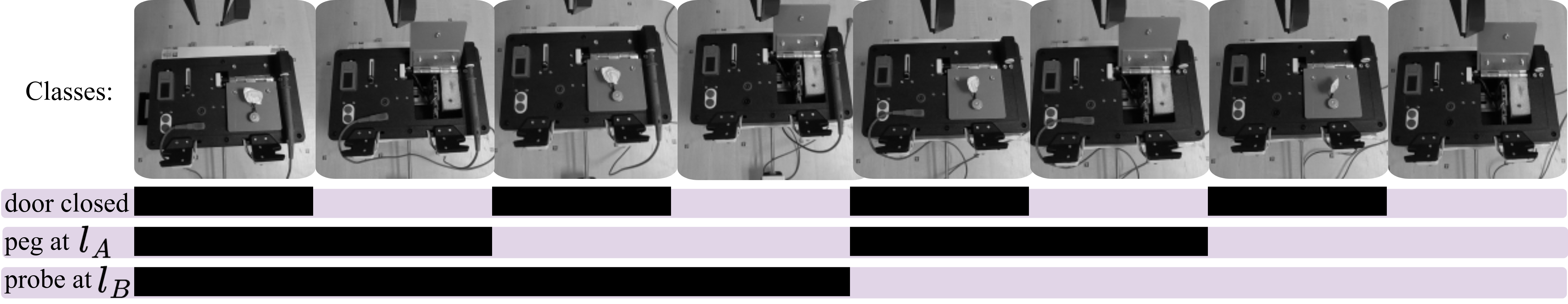}
    \vspace{-1.5em}
    \caption{\textbf{Model Reliability with Growing State Space.} Top: Model reliability under an increasing number of classes. As shown, the model \texttt{dinov2 small attn} performs well even with an increasing number of branches to choose from. 
    Bottom: Eight skill variants of the peg pick task. 
    }
    \vspace{-1.5em}
    \label{fig:growingclasses}
\end{figure}
We first evaluate Switcher scalability in a controlled experiment where the number of competing decision-state classes increases from 2 to 8. We vary three binary factors on the task-board: (i) door open/closed, (ii) peg at $l_A$/missing, and (iii) probe at $l_B$/missing, i.e., $8$ distinct state variations (Fig.~\ref{fig:growingclasses}, bottom). For each variation, we verify on five test executions. 
As baselines, we include SIFT matching and an AEGP model~(Sec.~\ref{sec:baselines}), which quickly degrade as class counts increase. As shown in Fig.~\ref{fig:growingclasses} (top), the proposed DINO-based attention model \texttt{dinov2 small attn} has the best decision success under growth of classes, remaining above $90\%$ up to 5 DS classes (compared to $62.5 \%$ for SIFT).

\begin{figure}
    \centering
    \vspace{0.5em}
    \includegraphics[width=0.99\linewidth]{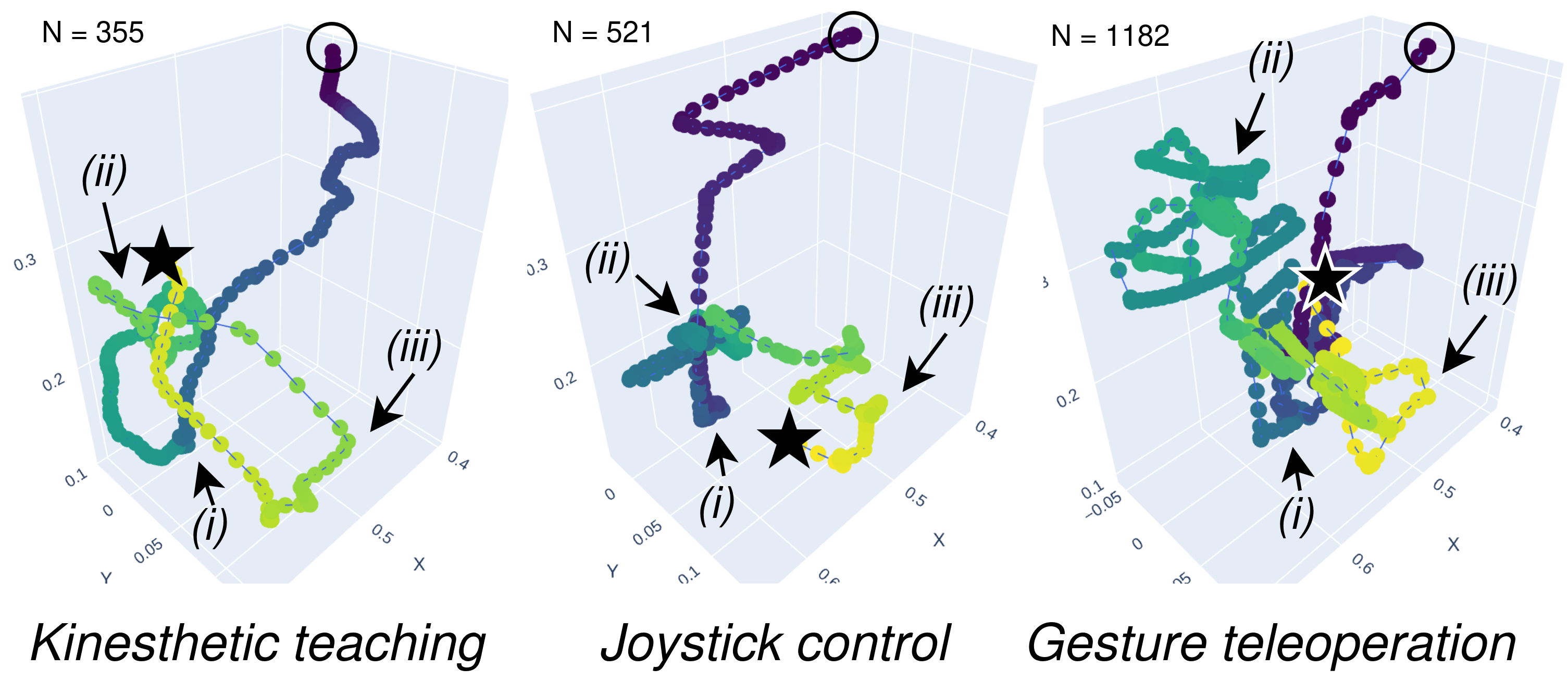}
    \vspace{-2em}
    \caption{\textbf{Example demonstration trajectories for \textit{Probe measure} task.} ${\color{black} \circ}$ is demo start. ${\color{black} \star}$~is demo end. \emph{(i)} is probe pick, \emph{(ii)} is measurement with probe, and \emph{(iii)} is probe drop into a bowl. You can see that gesture probe measurement \emph{(ii)} is messy when the user tries to reorient the gripper and focus on target precision. $N$ is the number of time steps. Keypoint colors indicate the time encoding. See video of recording at the link\textsuperscript{\protect\ref{websitefootnote}}.}
    \vspace{-2em}
    \label{fig:exampledemostrajectories}
\end{figure}

\vspace{-0.5em}
\subsection{Switcher evaluation on the novice-user dataset}
\label{sec:exp1}
We evaluate the Switcher on the user-study datasets (Sec.~\ref{sec:datasets}) using the \emph{Decision success} metric (Sec.~\ref{sec:replay_success_metric}). Each DS contains at most three candidate skill variants. To evaluate few-shot data aggregation, we move one or two additional rollouts per class into the training set and evaluate on the remaining held-out rollouts. No training rollout is used for testing.

\begin{figure}
    \centering
    \includegraphics[width=0.99\linewidth,trim={0.0cm 0.7cm 0.0cm 0.0cm},clip]{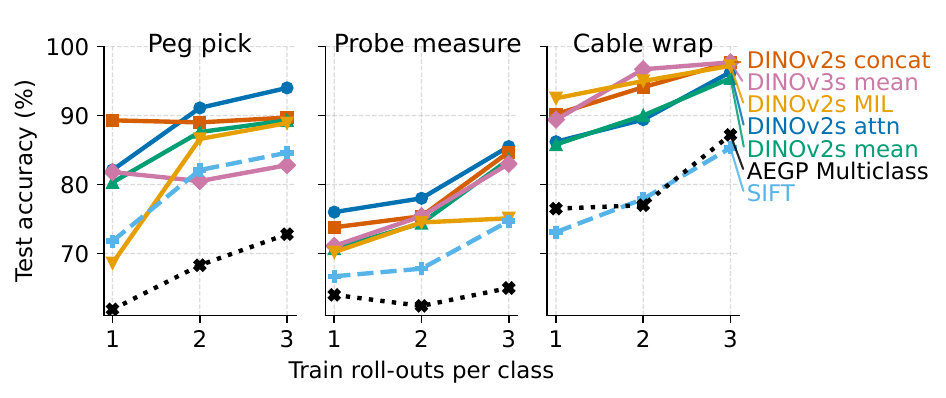}
    \includegraphics[width=0.99\linewidth,trim={0.0cm 0.0cm 0.0cm 0.7cm},clip]{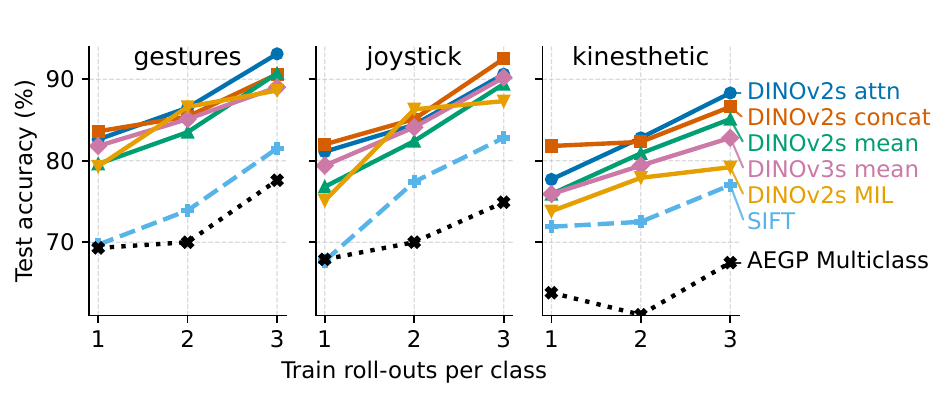}
    \vspace{-2.5em}
    \caption{\textbf{The evaluation of Switcher for considered methods} (Sec.~\ref{sec:switcher_methods}) on the collected task dataset $\mathcal{D}^{79}_\text{classification}$ averaged across evaluation on 79 small datasets.
    Modalities include \texttt{kin} for kinesthetic teaching, \texttt{joy} for joystick, and \texttt{gst} for hand-guided teleoperation.
    All methods scored min. 98\% accuracy on the training data.}
    \vspace{-1.5em}
    \label{fig:switcher_methods}
\end{figure}

\newcommand{\rotlabel}[1]{%
  \raisebox{1.4\height}{\rotatebox[origin=c]{90}{{\scriptsize \texttt{#1}}}}%
}

\begin{figure*}
    \centering
    \vspace{0.5em}
    \label{fig:exp1_results}
    \includegraphics[height=2.7cm]{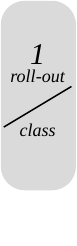}
    \includegraphics[width=0.26\linewidth,trim={0.3cm 0.3cm 0.45cm 0.3cm},clip]{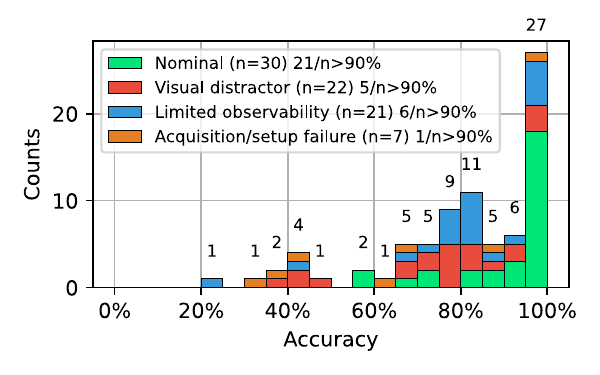} 
    \includegraphics[height=2.7cm]{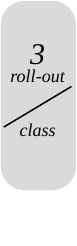}
    \includegraphics[width=0.26\linewidth,trim={0.3cm 0.3cm 0.45cm 0.3cm},clip]{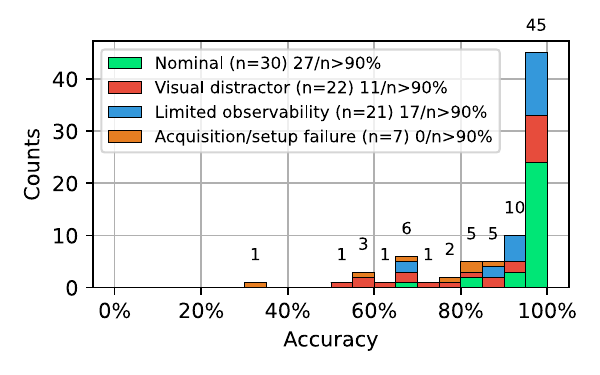}
    \includegraphics[height=2.7cm]{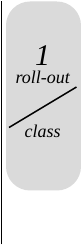}
    \includegraphics[width=0.26\linewidth,trim={0.3cm 0.3cm 0.3cm 0.3cm},clip]{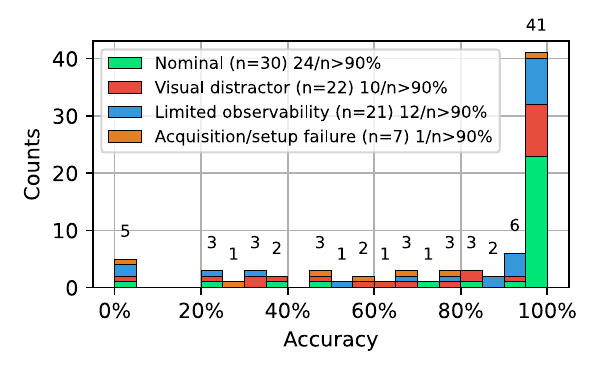} \\[-0.7em]
    \caption{\textbf{Per-decision-state (DS) branch-selection (left/center) and anomaly-detection (right) test accuracy distribution.} All results use the \texttt{dinov2 small attn} Switcher and are computed independently for each valid DS window. For branch selection, each DS window defines a multiclass branch-selection problem over the permitted successor skill parts. The branch-selection model is trained on images from the corresponding DS window using either one training rollout per class (left) or three training rollouts per class (center), and is evaluated on the remaining held-out rollouts. Thus, each DS window produces one branch-selection test accuracy. For anomaly detection, the same DS windows are relabeled as an in-distribution/OOD problem. For each DS window, images from the root skill part define the in-distribution training data, whereas images from all non-root skill parts are treated as out-of-distribution (OOD) samples. The anomaly detector is trained only on in-distribution images from the single rollout and is evaluated on all remaining held-out rollouts. The anomaly threshold is set with \texttt{percentile\_keep} $=0.1$. Each histogram aggregates the resulting per-DS test accuracies; bins denote accuracy intervals, numbers above bars denote the number of DS windows in each bin, and the \textit{y}-axis counts DS windows. 
    }
    \vspace{-1.4em}
    \label{fig:stateest_hist}
\end{figure*}

\subsubsection{Branch selection}
Fig.~\ref{fig:switcher_methods} reports multiclass DS classification accuracy across tasks and modalities. Frozen DINO features with lightweight classification heads perform reliably on the noisy user-study data. In the one-demonstration setting, \texttt{dinov2 small concat} achieves $82.38\%$ and \texttt{dinov2 small attn} achieves $80.70\%$. With two additional training rollouts per class, accuracy increases to $89.34\%$ and $90.62\%$, respectively.
Splitting the DS windows by acquisition quality (Fig.~\ref{fig:stateest_hist}) (per-category re-clustering: $30$ nominal / $22$ visual-distractor / $21$ limited-observability windows) localizes the failure modes for \texttt{dinov2 small attn}. With one training rollout per class, mean per-DS accuracy is $91.1\%$ nominal, $75.4\%$ under distractors, and $79.8\%$ under limited observability. Additional rollouts recover the limited-observability windows (to $93.2\%$) more readily than the distractor windows (to $84.3\%$), while nominal windows reach $96.7\%$.

\subsubsection{Anomaly detection}
Fig.~\ref{fig:stateest_hist} (right) summarizes anomaly-detection accuracy for the best-performing \texttt{dinov2 small attn} model. Both \texttt{dinov2 small concat} and \texttt{dinov2 small attn} are robust across DS instances, achieving $>90\%$ accuracy in $47/79$ and $46/79$ DS windows, respectively. 
\vspace{-1em}

\subsection{Novice user study: teaching efficiency and task success}
\label{sec:exp_userstudy}

\begin{table}[]
    \centering
    \vspace{1em}
    \begin{tabular}{lccc}
        Modality & \textit{Peg Pick} & \textit{Probe measure} & \textit{Cable wrap} \\
        \toprule
        \multicolumn{4}{c}{\textit{Task success (Estimated autonomous success) [\%]}} \\
        \midrule
        \texttt{kin} & 92.9 (81.4) & 93.3 (\textbf{76.2}) & 70.3 (\textbf{62.4}) \\
        \texttt{joy} & 88.2 (\textbf{84.7}) & 76.0 (56.3) & 60.0 (51.3) \\
        \texttt{gst} & 87.5 (83.3) & 84.5 (69.6) & 69.2 (61.2) \\
        \midrule
        \multicolumn{4}{c}{\textit{Demonstration length [s], mean $\pm$ std}} \\
        \midrule
        \texttt{kin} & \textbf{17.4 $\pm$ 4.2} & \textbf{19.5 $\pm$ 3.7} & \textbf{24.0 $\pm$ 5.4} \\
        \texttt{joy} & 26.1 $\pm$ 9.2 & 39.2 $\pm$ 18.6 & 65.3 $\pm$ 26.2 \\ 
        \texttt{gst} & 38.2 $\pm$ 28.7 &  41.6 $\pm$ 13.0 & 74.8 $\pm$ 38.1  \\
        \midrule
    \end{tabular}
    \caption{\textbf{User study results.}}
    \vspace{-3em}
    \label{tab:userstudy}
\end{table}

Task success (Table~\ref{tab:userstudy}, top) measures task execution under manual switching, isolating trajectory quality from perception.
Values in parentheses denote estimated \textit{autonomous success}, which links task replay success with the branch classifier evaluated offline (Sec.~\ref{sec:exp1}), computed as:
\begin{equation}
\label{eq:eas}
\widehat{S}_{\mathrm{auto}}(\tau,q)=
\frac{1}{N_{\tau,q}}
\sum_{r=1}^{N_{\tau,q}}
y_r \prod_{d \in D_r} \hat{a}_{r,d},
\end{equation}
where $\tau$ is the task, $q$ is the modality, $N_{\tau,q}$ is the number of evaluated rollouts, $y_r \in \{0,1\}$ is the task success for rollout $r$, $D_r$ are its traversed decision states, and $\hat{a}_{r,d} \in [0,1]$ the fraction of frames in DS window that the held-out \texttt{dinov2 small attn} branch classifier assigns to the correct successor.
Task success is highest for \emph{Peg pick} and \emph{Probe measure}, whereas \emph{Cable wrap} remains the most difficult task across modalities. Estimated autonomous success follows the same trend but decreases when branch selection is visually ambiguous, especially under limited observability. In these cases, the discriminative scene state, such as whether the probe door is open or closed, is not visible to the camera.

Demonstration length (Table~\ref{tab:userstudy}, bottom) indicates teaching effort. Kinesthetic teaching produced clearly the shortest demonstrations in all tasks. Teleoperation trajectories are longer  and also less smooth: mean path length grows from $1.7$\,m (kinesthetic) to $2.3$\,m (joystick) to $2.5$\,m (gesture), while spectral arc length degrades from $-7.7$ to $-17.6$ and $-19.0$, respectively, illustrated in Fig.~\ref{fig:exampledemostrajectories} for the probe task.

Despite these objective advantages of the kinesthetic teaching, all 8 participants (7 novices and 1 experienced) (within-subjects, a fixed modality order (kinesthetic, joystick, gesture)) report that they favor the teleoperation modalities. 

\vspace{-0.4em}
\section{Conclusion and Discussion}
\label{sec:discussion_usergated}

We presented \emph{See and Switch}, a conditional PbD framework that represents tasks as branching skill parts and extends them incrementally through decision states. The vision-based Switcher selects the appropriate successor skill part at each DS or flags observations outside the demonstrated contexts as anomalies.

\paragraph{Switcher performance and practical implications}
The results show that DS-local visual switching is feasible on noisy novice-user demonstrations. In the one-demonstration setting, \texttt{dinov2 small attn} achieves $80.70\%$ branch-selection accuracy; with two additional rollouts per class, accuracy increases to $90.62\%$. This supports the main premise of the framework: users can start from sparse demonstrations and improve autonomous branching through incremental rollout data. The controlled label-growth experiment further indicates that the same model remains reliable as the number of competing branches increases.

\paragraph{User study findings}
The user study shows that non-expert participants can teach conditional manipulation tasks through kinesthetic teaching, joystick control, and hand gestures. Kinesthetic teaching produced substantially shorter and smoother demonstrations, while gesture and joystick interfaces remained viable alternatives when direct physical guidance is unavailable. 
 The estimate of autonomous execution success (Eq.~\eqref{eq:eas}) in Table~\ref{tab:userstudy} gives a more complete view than task success alone.

\paragraph{Limitations and open problems}
The main limitation is observability, quantified by the quality split in Sec.~\ref{sec:exp1}: switching degrades when the relevant scene information is outside the eye-in-hand camera view or occluded by distractors, motivating multi-view sensing and active viewpoint selection. Another limitation is the current lack of mechanisms for branch rejoining after the decisive task variation has been resolved.
\vspace{-1em}

\bibliographystyle{IEEEtran}
\bibliography{root}

\end{document}